\documentclass[11pt]{article}
\usepackage[preprint]{acl}
\usepackage{times}
\usepackage{latexsym}
\usepackage{graphicx}
\usepackage{booktabs}
\usepackage{amsmath}
\usepackage{amssymb}
\usepackage{multirow}
\usepackage{xcolor}
\definecolor{palWrite}{HTML}{E07B54}
\definecolor{palRetr}{HTML}{5185A0}
\definecolor{palEPC}{HTML}{2AA876}
\definecolor{palNeutral}{HTML}{8C8C8C}
\definecolor{palDark}{HTML}{2C3E50}
\definecolor{palLightW}{HTML}{F2D0C2}
\definecolor{palLightR}{HTML}{C8D9E6}
\definecolor{palLightE}{HTML}{C2E8D8}
\definecolor{palProbe}{HTML}{1E9C89}
\definecolor{palEvidence}{HTML}{4D89A8}
\definecolor{palSelect}{HTML}{D96B4D}
\definecolor{palBudget}{HTML}{D6A64A}
\definecolor{palNote}{HTML}{F7F8FA}
\usepackage{tikz}
\usetikzlibrary{arrows.meta, positioning, shapes.geometric, calc, decorations.pathreplacing, fit, backgrounds}
\usepackage{subcaption}
\usepackage[T1]{fontenc}
\usepackage[utf8]{inputenc}
\usepackage{microtype}
\DeclareMathOperator*{\argmin}{arg\,min}

\title{WhenLoss: Diagnosing Write and Retrieval Bottlenecks\\in Long-Context Memory Systems}

\author{
Jiangnan Yu, Kisson Songqi Lin, and Jilong Wu
}

\begin{document}
\maketitle

\begin{abstract}
Long-context memory systems often fail under fixed budgets, but end-to-end evaluation does not reveal whether evidence was discarded during compression or preserved but never retrieved.
We introduce a \textbf{four-condition diagnostic protocol} that evaluates a fixed reader under truncated full context (\textsc{TFC}), oracle evidence (\textsc{OE}), complete stored memory (\textsc{CSM}), and retrieved memory (\textsc{RM}).
Under this fixed-budget LongMemEval setup, write-side gaps exceed retrieval-side gaps for most tested baselines, with four of six baselines robustly \emph{write-dominant} under our default diagnosis margin.
Motivated by this diagnosis, we propose \textbf{Expected Predictive Compression (EPC)}, which moves the key decision---what information to retain---to write time by using an LLM to anticipate likely future questions and preserve the minimal supporting evidence under the token budget, while leaving retrieval unchanged at question time.
Across all 500 LongMemEval questions with three readers (GPT-5.2, Claude Sonnet~4, Gemini 2.5 Pro), EPC achieves the highest \textsc{CSM} scores among all systems (0.49 vs.\ 0.44 for Summary (LLM), the strongest baseline), reducing $\Delta_{\text{write}}$ to 0.04 while leaving $\Delta_{\text{retr}}$ comparable to other LLM-based systems.
These results suggest that, on this benchmark and evaluation setup, improving what the write stage preserves is a key avenue for performance gains in the tested systems.
\end{abstract}

\section{Introduction}

A user chats with an AI assistant for three months---sharing restaurant preferences, travel plans, project deadlines.
One day they ask: ``What Thai place did I mention liking in April?''
The system fails.
But \emph{why} it fails matters: did the memory system discard that detail during compression, or is the detail stored in memory but missed by retrieval?
The first case calls for better compression; the second, better retrieval.
End-to-end accuracy---the only metric most benchmarks report---cannot distinguish the two.

This is not a hypothetical problem.
Memory-augmented LLMs now use a wide range of write strategies---chunked storage \citep{lewis2020retrieval}, session summarization \citep{wang2024augmenting}, summary trees \citep{chen2024walking}, gist pages \citep{lee2024human}, importance-scored banks \citep{zhong2024memorybank}---but evaluations usually report only end-to-end performance.
When a system scores poorly, practitioners cannot tell whether to invest in compression or retrieval.

We expose these stage-level effects by evaluating the same reader under four controlled inputs that replace or bypass different parts of the memory pipeline.
The resulting \textbf{four-condition diagnostic protocol} (\S\ref{sec:protocol}) compares oracle evidence (\textsc{OE}) against complete stored memory (\textsc{CSM}) to estimate the write-side gap, and \textsc{CSM} against retrieved memory (\textsc{RM}) for the retrieval-side gap.
The protocol is operational---it localizes where performance degrades, without claiming to identify root causes.

Applying this protocol to six baseline systems on LongMemEval \citep{wu2025longmemeval}, we find a consistent asymmetry: write-side gaps exceed retrieval-side gaps for most baselines, with four baselines robustly write-dominant under our default diagnosis margin.
For these baseline operating points, more performance is lost during the write stage than during retrieval.

This finding motivates using anticipated future questions to guide compression at write time.
The challenge is that compression happens \emph{before} the question arrives.
A standard summarizer does not know which facts will matter later.
\textbf{Expected Predictive Compression (EPC)} (\S\ref{sec:epc}) addresses this by prompting the LLM to generate likely future questions about the conversation.
Those self-generated probe questions then guide evidence selection under a fixed token budget---preserving the specific facts, dates, and preferences most likely to be needed, rather than producing a generically readable summary.

Across 500 LongMemEval questions with three readers (GPT-5.2, Claude Sonnet~4, Gemini 2.5~Pro), EPC achieves the highest \textsc{CSM} score among all tested systems, with the lowest write-side gap ($\Delta_{\text{write}} = 0.04$).
On a second benchmark (LoCoMo), EPC again has the lowest write-side gap ($\Delta_{\text{write}} = 0.06$).
A cost-matched comparison separates the gain from an additional LLM call, and a budget sweep shows the advantage is largest under tight budgets---precisely where choosing the right evidence matters most.

\section{Related Work}

\paragraph{Long-Context Memory Systems.}
Prior work explores a wide range of memory systems for long interactions.
MemGPT \citep{packer2024memgpt} introduces a virtual memory hierarchy; MemWalker \citep{chen2024walking} organizes long documents as a summary tree; ReadAgent \citep{lee2024human} stores gists and expands them at read time; MemoryBank \citep{zhong2024memorybank} scores memory by importance and temporal decay; and \citet{wang2024augmenting} train a long-term memory module jointly with the model.
\citet{zhang2024survey} survey these memory mechanisms more broadly.
Our focus is complementary: rather than proposing another end-to-end system, we ask how to \emph{localize where performance degrades} in a memory pipeline.

\paragraph{Memory Evaluation and Error Analysis.}
LongMemEval \citep{wu2025longmemeval} provides multi-session conversations with gold turn-level evidence annotations, making it possible to evaluate memory systems against explicit supporting evidence.
In retrieval-augmented generation, \citet{xu2024retrieval} compare retrieval-based and long-context approaches, but they evaluate end-to-end performance without separating write-side from retrieval-side degradation.
Our protocol fills that gap.

\paragraph{Compression for Downstream Tasks.}
Prompt and context compression methods \citep{jiang2023llmlingua,pan2024llmlingua2} reduce token counts while preserving general utility.
Query-aware variants---RECOMP \citep{xu2023recomp} and the broader query-focused summarization tradition \citep{daume2006bayesian,baumel2018query}---go further by conditioning compression on a known downstream query.
EPC instead operates before the downstream query is available.
It approximates the future-question distribution via LLM self-questioning and minimizes expected answer loss over likely future questions rather than optimizing generic readability or a response to a known query.

\section{Diagnostic Protocol}
\label{sec:protocol}

This section defines the four controlled input conditions, the resulting write-side and retrieval-side indicators, and the rule used to label the dominant bottleneck.

\subsection{Problem Formulation}

Consider a memory-augmented QA system operating over a conversation history $H = \{s_1, s_2, \ldots, s_n\}$ consisting of $n$ sessions.
Given a question $q$ with gold answer $y$ and gold evidence turns $E_q$ drawn from $H$, the system must:
(1) \textbf{write} the history into a budget-limited memory store $M$ with capacity $B$ tokens, and
(2) \textbf{read} from $M$ by retrieving content $R \subseteq M$ to provide context for answering $q$.
Here, the write stage denotes the pre-question processing that stores, indexes, evicts, or compresses history into $M$; retrieval is the question-time selection of content from that store.

\subsection{Four Conditions}

\begin{figure}[t]
\centering
\pgfdeclarelayer{bg}
\pgfsetlayers{bg,main}
\resizebox{\columnwidth}{!}{%
\begin{tikzpicture}[
    node distance=0.5cm,
    pstage/.style={
        rectangle, rounded corners=5pt, minimum width=2.0cm, minimum height=0.72cm,
        align=center, inner sep=5pt, line join=round, font=\small\bfseries,
        draw=#1!80!palDark, fill=#1!28, line width=1.3pt,
    },
    condbox/.style={
        rectangle, rounded corners=4pt, minimum width=1.55cm, minimum height=0.6cm,
        align=center, inner sep=4pt, line join=round, font=\footnotesize\bfseries,
        draw=#1!55!palDark, fill=#1!12, line width=0.9pt,
    },
    readerbox/.style={
        rectangle, rounded corners=4pt, minimum width=1.3cm, minimum height=0.6cm,
        align=center, inner sep=4pt, font=\footnotesize\bfseries,
        draw=palNeutral!45, fill=white, line width=0.9pt,
    },
    flow/.style={-{Stealth[length=3.0mm,width=2.4mm]}, line width=1.15pt, draw=palDark!75},
    feed/.style={-{Stealth[length=2.2mm,width=1.8mm]}, line width=0.9pt, draw=palDark!60},
    pipeband/.style={
        rectangle, rounded corners=7pt, draw=palWrite!25, fill=palWrite!8,
        line width=0.6pt, inner sep=9pt,
    },
    condband/.style={
        rectangle, rounded corners=7pt, draw=palNeutral!25, fill=palNote,
        line width=0.6pt, inner sep=7pt,
    },
]
    \node[pstage=palWrite] (H) at (0,0) {$H$};
    \node[pstage=palWrite] (M) at (3.5,0) {$M$};
    \node[pstage=palRetr]  (R) at (7.0,0) {$R$};

    \draw[flow] (H) -- node[above=3pt, font=\small\bfseries, text=palDark!85] {write} (M);
    \draw[flow] (M) -- node[above=3pt, font=\small\bfseries, text=palDark!85] {retrieve} (R);

    \node[condbox=palEPC, minimum width=1.0cm, font=\small\bfseries] (E) at (1.75,-0.9) {$E_q$};
    \draw[feed, densely dashed, draw=palEPC!75]
        (H.south east) -- (E.north west)
        node[midway, above right=0pt and 1pt, font=\scriptsize, text=palDark!55] {gold};

    \node[condbox=palNeutral] (FC)  at (0,   -2.3) {\textsc{TFC}};
    \node[condbox=palEPC]     (OE)  at (2.35,-2.3) {\textsc{OE}};
    \node[condbox=palWrite]   (CSM) at (4.7, -2.3) {\textsc{CSM}};
    \node[condbox=palRetr]    (RM)  at (7.0, -2.3) {\textsc{RM}};

    \draw[feed] (H.south)  -- (FC.north);
    \draw[feed] (E.south)  -- (OE.north);
    \draw[feed] (M.south)  -- (CSM.north);
    \draw[feed] (R.south)  -- (RM.north);

    \begin{pgfonlayer}{bg}
        \node[pipeband, fit=(H) (M) (R)] {};
        \node[condband, fit=(FC) (OE) (CSM) (RM)] (condset) {};
    \end{pgfonlayer}

    \node[readerbox] (reader) at (9.0,-2.3) {Reader};
    \draw[flow, draw=palDark!55] (condset.east) -- node[above=2pt, font=\scriptsize, text=palDark!60] {one at a time} (reader.west);
    \node[font=\normalsize\bfseries, text=palDark] (score) at (10.2,-2.3) {$\phi$};
    \draw[flow] (reader) -- (score);

    \draw[decorate, decoration={brace, amplitude=5pt, mirror}, line width=1.1pt, palWrite]
        ([yshift=-10pt]OE.south west) -- ([yshift=-10pt]CSM.south east)
        node[midway, below=7pt, font=\small\bfseries, palWrite] {$\Delta_{\text{write}}$};
    \draw[decorate, decoration={brace, amplitude=5pt, mirror}, line width=1.1pt, palRetr]
        ([yshift=-10pt, xshift=1pt]CSM.south east) -- ([yshift=-10pt]RM.south east)
        node[midway, below=7pt, font=\small\bfseries, palRetr] {$\Delta_{\text{retr}}$};
\end{tikzpicture}%
}
\caption{The four-condition diagnostic protocol. Each condition is evaluated by the same reader; $\Delta_{\text{write}} = \phi(\textsc{OE}) - \phi(\textsc{CSM})$ and $\Delta_{\text{retr}} = \phi(\textsc{CSM}) - \phi(\textsc{RM})$.}
\label{fig:protocol}
\end{figure}
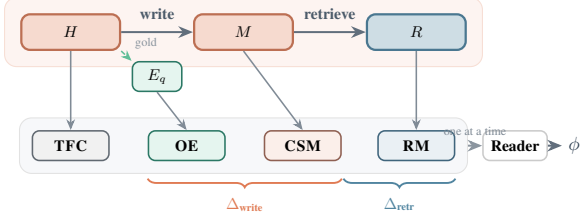

We define four evaluation conditions, each providing a different input to a fixed reader, to isolate reference performance and memory-pipeline effects (Figure~\ref{fig:protocol}):

\begin{description}
    \item[\textsc{TFC} (Truncated Full Context):] The reader receives $H$ truncated to a fixed 32K-token context budget. No memory system is involved.
    \item[\textsc{OE} (Oracle Evidence):] The reader receives only the gold evidence turns $E_q$---distractors removed, no truncation.
    \item[\textsc{CSM} (Complete Stored Memory):] The reader receives all content in memory $M$ after the write stage.
    \item[\textsc{RM} (Retrieved Memory):] The reader receives the retrieved subset $R \subseteq M$.
\end{description}

\noindent Two gaps localize where performance drops within the memory pipeline: \textsc{OE}$\to$\textsc{CSM} defines the write-side gap, and \textsc{CSM}$\to$\textsc{RM} defines the retrieval-side gap.
Two caveats apply.
First, the \textsc{OE}--\textsc{TFC} gap mixes distractor removal with the imposed context budget, so \textsc{OE} $>$ \textsc{TFC} does not isolate either effect.
Second, the \textsc{OE}$\to$\textsc{CSM} gap should be read as an upper bound on write-side degradation: it can also reflect format mismatch between stored memory and reader expectations, or loss of contextual cues.
These gaps are therefore \emph{operational} indicators that localize where performance degrades, not causal attributions to a single mechanism.
While each condition is individually simple, prior evaluations report only end-to-end scores (\textsc{RM} or equivalent), making it impossible to determine whether failures originate in compression or retrieval.
The protocol's contribution is combining these conditions into a reusable diagnostic that any memory system can run without architectural changes.

\subsection{Bottleneck Indicators}

For a scoring metric $\phi$, we use the two gaps as \emph{operational bottleneck indicators}:
\begin{align}
    \Delta_{\text{write}} &= \phi(\textsc{OE}) - \phi(\textsc{CSM}) \label{eq:write_loss} \\
    \Delta_{\text{retr}} &= \phi(\textsc{CSM}) - \phi(\textsc{RM}) \label{eq:retrieval_loss}
\end{align}
These are additive: $\phi(\textsc{OE}) - \phi(\textsc{RM}) = \Delta_{\text{write}} + \Delta_{\text{retr}}$.
For compact notation, we write $\Delta_w=\Delta_{\text{write}}$ and $\Delta_r=\Delta_{\text{retr}}$ below.

\subsection{Diagnosis Rule}

With margin $\epsilon = 0.02$:
\begin{equation}
\text{diagnosis} =
\begin{cases}
    \textsc{Write} & \text{if } \Delta_w > \Delta_r + \epsilon \\
    \textsc{Retrieval} & \text{if } \Delta_r > \Delta_w + \epsilon \\
    \textsc{Mixed} & \text{otherwise}
\end{cases}
\end{equation}
The margin only affects near-balanced systems; four baselines remain unambiguously \textsc{Write} under $\epsilon \in [0,.05]$, while Summary (LLM) and ReadAgent lie near the boundary.

\section{Expected Predictive Compression}
\label{sec:epc}

Consider LLM summarization, the strongest query-agnostic compression baseline among our tested systems (Table~\ref{tab:main_diag}).
It compresses 121K tokens of conversation into 5K tokens of fluent text---yet even when the reader has access to all stored memory, fewer than half the questions are answered correctly (CSM = .44).
Which information is lost?
The summarizer preserves broadly salient content---topics discussed, decisions made---but omits specific dates, entity names, and preference details that downstream questions target.
This motivates a compression objective that prioritizes supporting evidence, not just readable summaries.

We propose \textbf{Expected Predictive Compression (EPC)}, which shifts the compression objective accordingly.

\subsection{Formulation}

Given a conversation segment $x$ (a session in our experiments) and a token budget $B$, let $Q(x)$ denote a set of possible future questions that may require information from $x$, with $w(q)$ as each question's estimated likelihood.
Let $A(c, q)$ denote the reader's answer to question $q$ given context $c$, and $\mathcal{L}$ a loss function measuring answer degradation.
EPC seeks the compressed memory $m^*$ that minimizes expected answer loss:

\begin{equation}
    m^* = \argmin_{|m| \leq B} \sum_{q \in Q(x)} w(q) \cdot \mathcal{L}\big(A(x, q),\; A(m, q)\big)
    \label{eq:epc}
\end{equation}

where the weighted sum approximates the expectation over the future-question distribution.
This is loosely motivated by rate--distortion theory \citep{cover2006elements}, where the budget $B$ plays the role of a rate constraint and the distortion is future-question answer loss rather than surface-level reconstruction error.
The actual implementation uses a greedy heuristic (Eq.~\ref{eq:utility}) rather than formal rate--distortion optimization; the connection is conceptual rather than algorithmic.

\subsection{LLM Self-Questioning EPC}

Since $Q(x)$ is unknown at write time, we approximate it via \textbf{LLM self-questioning}: the LLM first generates probe questions that the segment is likely to be queried about, then uses them to guide evidence selection.
The implementation treats those generated probe questions as an unweighted approximation to $Q(x)$.
Figure~\ref{fig:epc} illustrates the procedure:

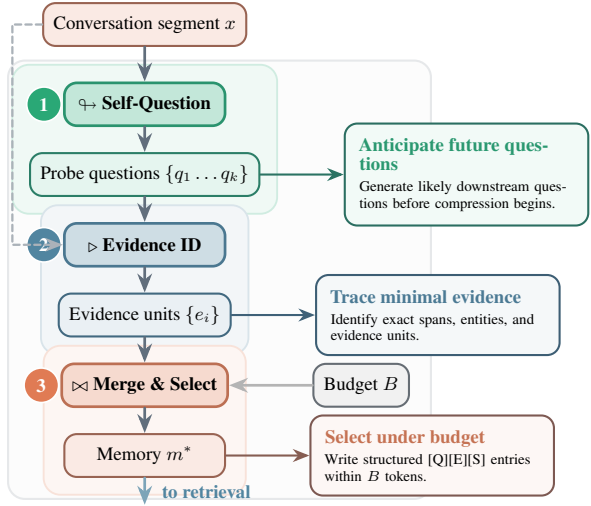
\begin{figure}[t]
\centering
\pgfdeclarelayer{background}
\pgfsetlayers{background,main}
\resizebox{\columnwidth}{!}{%
\begin{tikzpicture}[
    node distance=0.42cm,
    basebox/.style={
        rectangle, rounded corners=6pt, minimum width=2.7cm, minimum height=0.72cm,
        align=center, inner sep=5pt, line join=round,
    },
    artifact/.style={
        basebox, draw=#1!60!palDark, fill=#1!12, line width=1.0pt, font=\footnotesize,
    },
    stage/.style={
        basebox, draw=#1!80!palDark, fill=#1!28, line width=1.3pt, font=\footnotesize\bfseries,
    },
    badge/.style={
        circle, fill=#1, draw=white, text=white, minimum size=5.8mm,
        inner sep=0pt, font=\footnotesize\bfseries, line width=0.8pt,
    },
    notecard/.style={
        rectangle, rounded corners=6pt,
        draw=#1!45!palDark, fill=#1!6,
        text width=3.6cm, minimum height=1.05cm, align=left,
        inner sep=7pt, line width=1.0pt, font=\scriptsize, line join=round,
    },
    flow/.style={-{Stealth[length=3.2mm,width=2.5mm]}, line width=1.3pt, draw=palDark!75, line cap=round},
    auxflow/.style={-{Stealth[length=3.0mm,width=2.3mm]}, line width=1.1pt, draw=palDark!40, densely dashed, line cap=round},
    noteedge/.style={-{Stealth[length=2.8mm,width=2.2mm]}, line width=1.0pt, draw=#1!50!palDark, line cap=round},
    stageband/.style={
        rectangle, rounded corners=8pt, draw=#1!22, fill=#1!7, line width=0.9pt, inner sep=8pt,
    },
    pipebg/.style={
        rectangle, rounded corners=8pt, draw=palDark!12, fill=palNeutral!3,
        line width=1.0pt, inner sep=10pt,
    },
]
    \node[artifact=palWrite, fill=palWrite!16, minimum width=2.9cm] (chunk) {Conversation segment $x$};

    \node[stage=palEPC, below=0.58cm of chunk] (s1) {\raisebox{-0.5pt}{\small$\looparrowright$}~Self-Question};
    \node[badge=palEPC] at ([xshift=-0.32cm]s1.west) {1};
    \draw[flow] (chunk) -- (s1);

    \node[artifact=palEPC, below=of s1] (probes) {Probe questions $\{q_1 \ldots q_k\}$};
    \draw[flow] (s1) -- (probes);

    \node[stage=palRetr, below=of probes] (s2) {\raisebox{-0.5pt}{\small$\triangleright$}~Evidence ID};
    \node[badge=palRetr] at ([xshift=-0.32cm]s2.west) {2};
    \draw[flow] (probes) -- (s2);
    \draw[auxflow] (chunk.west) -- ++(-0.5,0) |- (s2.west);

    \node[artifact=palRetr, below=of s2] (evidence) {Evidence units $\{e_i\}$};
    \draw[flow] (s2) -- (evidence);

    \node[stage=palWrite, below=of evidence] (s34) {\raisebox{-0.5pt}{\small$\bowtie$}~Merge \& Select};
    \node[badge=palWrite] at ([xshift=-0.32cm]s34.west) {3};

    \node[artifact=palNeutral, right=1.4cm of s34, minimum width=1.6cm, fill=palNeutral!12] (budget) {Budget $B$};
    \draw[flow, draw=palNeutral!65] (budget) -- (s34);

    \draw[flow] (evidence) -- (s34);

    \node[artifact=palWrite, fill=palWrite!16, below=of s34] (memory) {Memory $m^*$};
    \draw[flow] (s34) -- (memory);
    \draw[flow, draw=palRetr!75] (memory.south) -- ++(0,-0.45)
        node[midway, right=0.14cm, font=\footnotesize\bfseries, text=palRetr] {to retrieval};

    \begin{pgfonlayer}{background}
        \node[pipebg, fit=(s1) (probes) (s2) (evidence) (s34) (budget) (memory)] {};
        \node[stageband=palEPC, fit=(s1) (probes)] {};
        \node[stageband=palRetr, fit=(s2) (evidence)] {};
        \node[stageband=palWrite, fit=(s34) (memory)] {};
    \end{pgfonlayer}

    \node[notecard=palEPC, right=1.4cm of probes] (n1) {
        {\small\bfseries\color{palEPC!85!palDark} Anticipate future questions}\\[2pt]
        Generate likely downstream questions before compression begins.
    };
    \draw[noteedge=palEPC] (probes.east) -- (n1.west);

    \node[notecard=palRetr, right=1.4cm of evidence] (n2) {
        {\small\bfseries\color{palRetr!85!palDark} Trace minimal evidence}\\[2pt]
        Identify exact spans, entities, and evidence units.
    };
    \draw[noteedge=palRetr] (evidence.east) -- (n2.west);

    \node[notecard=palWrite, right=1.4cm of memory] (n3) {
        {\small\bfseries\color{palWrite!85!palDark} Select under budget}\\[2pt]
        Write structured [Q][E][S] entries within $B$ tokens.
    };
    \draw[noteedge=palWrite] (memory.east) -- (n3.west);

\end{tikzpicture}%
}
\caption{The EPC write pipeline. \textcircled{\scriptsize 1}~Generate probe questions. \textcircled{\scriptsize 2}~Identify supporting evidence. \textcircled{\scriptsize 3}~Merge and select under budget $B$.}
\label{fig:epc}
\end{figure}

\paragraph{Step 1: Generate probe questions.}
Given segment $x$, prompt the LLM to generate $k=5$ likely future questions targeting factual details, preferences, plans, temporal information, and state changes.

\paragraph{Step 2: Identify supporting evidence.}
For each probe question $q_i$, the LLM identifies the minimal supporting evidence: specific turns, spans, and entities.

\paragraph{Step 3: Merge, score, and select.}
Overlapping evidence spans are merged, and each evidence unit $e$ receives a utility score:
\begin{equation}
\begin{aligned}
u(e) ={}& \alpha \cdot \text{coverage}(e) + \beta \cdot \text{specificity}(e) \\
&- \lambda \cdot \text{redundancy}(e)
\end{aligned}
\label{eq:utility}
\end{equation}
where coverage counts how many probe questions $e$ supports; specificity rewards named entities, dates, and numbers; and redundancy penalizes overlap with already-selected evidence.
At each greedy selection step, units are ranked by the current $u(e)$ until the budget $B$ is exhausted, and each selected unit is written as a structured memory entry:

\begin{small}
\begin{verbatim}
[Q] What food does the user prefer?
[E] User prefers Thai over Italian.
[S] session_12_turn_3
\end{verbatim}
\end{small}

\paragraph{Design rationale.}
The design assumes that LLM-generated probe questions can cover common future question targets such as preferences, facts, and temporal events, so even an approximate $Q(x)$ can direct compression toward useful evidence.
Unlike conventional summarization, which optimizes for readability, EPC prioritizes supporting evidence---producing structured, information-dense output that preserves exact entities, dates, and numbers.

\section{Experimental Setup}

This section fixes the dataset, memory systems, model configurations, retrieval setup, and metrics used for the main LongMemEval experiments.

\subsection{Dataset}
We evaluate on all 500 questions from LongMemEval \citep{wu2025longmemeval}.
The benchmark is well suited to our setting because it pairs multi-session conversations averaging 121K tokens (${\sim}$50 sessions per conversation) with gold turn-level evidence annotations.
We keep the dataset's original session boundaries throughout.

\subsection{Memory Systems}
Table~\ref{tab:systems} lists the seven memory systems we compare.
The baselines span raw storage, heuristic summarization, LLM-based compression, hierarchical memory, and importance-weighted memory to cover diverse write and retrieval designs.
For controlled comparison, EPC shares the same budgets and embedding top-$k$ retriever as the Summary systems and the same writer model as Summary (LLM).
Because \textsc{CSM} removes retrieval, EPC's \textsc{CSM} gains over Summary (LLM) reflect the write-side compression strategy rather than retriever choice.

\paragraph{Baseline provenance.}
Verbatim Chunk and the two Summary variants are implemented directly in our codebase.
MemWalker, ReadAgent, and MemoryBank were reimplemented following the procedures described in the original papers \citep{chen2024walking,lee2024human,zhong2024memorybank} and adapted to the session-based LongMemEval format: MemWalker constructs a summary tree over sessions using the same writer model (GPT-5.2); ReadAgent stores gist pages per session and expands them at read time; MemoryBank scores memories by importance with temporal decay.
None of these systems provide official LongMemEval evaluation code, so reimplementation differences may affect absolute scores; however, all systems use the same dataset, budget, and readers, so relative comparisons within our evaluation are controlled.
Full prompts, hyperparameters, baseline adaptation details, and significance-testing procedures are provided in Appendix~\ref{app:repro}.

\begin{table}[t]
\centering
\small
\resizebox{\columnwidth}{!}{%
\begin{tabular}{lll}
\toprule
\textbf{System} & \textbf{Write Strategy} & \textbf{Retrieval} \\
\midrule
Verbatim Chunk & Raw chunks, FIFO eviction & Emb.\ top-$k$ \\
Summ.\ (Extractive) & Heuristic extractive & Emb.\ top-$k$ \\
Summ.\ (LLM) & LLM compression (query-agnostic) & Emb.\ top-$k$ \\
\textbf{EPC (ours)} & \textbf{LLM self-question + evidence} & \textbf{Emb.\ top-$k$} \\
MemWalker & Summary tree & Tree nav. \\
ReadAgent & LLM gist pages & Lookup expansion \\
MemoryBank & Importance + temporal decay & Importance+emb. \\
\bottomrule
\end{tabular}
}
\caption{Seven memory systems. EPC shares the same budgets and embedding retriever with the Summary systems, and shares the writer model with Summary (LLM); \textsc{CSM} comparisons remove retrieval and isolate the write-side compression strategy.}
\label{tab:systems}
\end{table}

\subsection{Implementation Details}

\paragraph{Readers.}
We use three readers: GPT-5.2 (\texttt{gpt-5.2}), Claude Sonnet~4 (\texttt{claude-sonnet-4-20250514}), and Gemini 2.5~Pro (\texttt{gemini-2.5-pro}).
All readers use temperature 0, max output tokens 200, and a simple prompt: ``Based on the following context, answer the question. If the answer cannot be determined, say `I don't know'.''
The context field is populated with the truncated full history (\textsc{TFC}, fixed 32K-token budget), gold evidence turns (\textsc{OE}), complete stored memory (\textsc{CSM}), or retrieved memory (\textsc{RM}).
For \textsc{TFC}, truncation retains the \emph{most recent} 32K tokens (tail of the conversation), discarding earlier sessions.
This is a budget-limited long-context baseline, not an unconstrained full-context reference.

\paragraph{Memory budgets and chunking.}
Write budget $B$=5{,}000 tokens; read budget 5{,}000 tokens.
Token counts use the \texttt{cl100k\_base} tokenizer.
For Verbatim Chunk, conversations are split into 200-token chunks with no overlap; FIFO eviction drops the oldest chunks when exceeding $B$.
For Summary systems and EPC, each session is compressed independently; per-session budgets are allocated proportionally to the square root of session length.

\paragraph{Retrieval.}
Verbatim Chunk, Summary (Extractive), Summary (LLM), and EPC all use the same retriever: \texttt{all-MiniLM-L6-v2} sentence embeddings \citep{reimers2019sentence,wang2020minilm} with cosine similarity, returning top-$k$=5 chunks/entries.
MemWalker uses its own tree navigation; ReadAgent uses lookup expansion; MemoryBank uses importance-weighted re-ranking over the same embedding scores.

\paragraph{Write-side LLM configuration.}
The directly compared LLM compression methods---Summary (LLM), EPC, and the 2-pass baseline---use the same model (GPT-5.2, temperature 0) for compression, ensuring that CSM differences among them reflect the compression \emph{strategy}, not the model.
Summary (LLM) uses a single call per session instructing the model to summarize under the token budget while preserving names, dates, numbers, preferences, and decisions, without guessing future questions.
The 2-pass baseline uses the same first pass, then a second call asking the model to improve information density by replacing vague references with specific facts.
EPC uses two calls: Step~1 generates 5 probe questions; Step~2 identifies supporting evidence (see \S\ref{sec:epc}).
Utility weights in Eq.~\ref{eq:utility}: $\alpha$=1.0, $\beta$=0.5, $\lambda$=0.3, selected in a 20-example pilot before the final evaluation.
Full prompt templates for all three methods are provided in Appendix~\ref{app:repro}.

\subsection{Metrics}
We report \textbf{Contains Match} (CM): whether the generated answer contains the gold answer substring.
We also report \textbf{Token F1}: token-level precision/recall between generated and gold answers.
Both are standard LongMemEval metrics \citep{wu2025longmemeval}.
All scores are averaged over 500 questions.
For the 3-reader tables, we report the mean across readers.
\textsc{TFC} and \textsc{OE} are shared reference conditions, computed once per reader and reused across all systems.

CM is a strict substring test; we verify in Appendix~\ref{app:repro} that CM and F1 agree directionally in 95.5\% of pairwise comparisons, and system rankings are identical under both metrics in all tables.

\section{Results}

We first verify that the reference conditions make the diagnostic meaningful, then report the main write--retrieval decomposition.
The remaining subsections test whether the diagnosis is supported by reader-independent evidence preservation, controlled perturbations, component ablations, cost matching, cross-benchmark generalization, and budget sensitivity.

\subsection{Reader-Level Reference Conditions}

\begin{table}[t]
\centering
\small
\setlength{\tabcolsep}{2pt}
\begin{tabular}{lcccc}
\toprule
\textbf{Reader} & \textbf{TFC\,CM} & \textbf{OE\,CM} & \textbf{TFC\,F1} & \textbf{OE\,F1} \\
\midrule
GPT-5.2 & 0.18 & 0.55 & 0.29 & 0.65 \\
Claude S4 & 0.17 & 0.50 & 0.28 & 0.61 \\
Gemini 2.5 & 0.19 & 0.53 & 0.30 & 0.63 \\
\midrule
\textbf{Mean} & \textbf{.18} & \textbf{.53} & \textbf{.29} & \textbf{.63} \\
\bottomrule
\end{tabular}
\caption{Reader-level reference conditions on LongMemEval. \textsc{OE} $\gg$ \textsc{TFC} across all readers.}
\label{tab:reader_refs}
\end{table}

Table~\ref{tab:reader_refs} shows that all three readers exhibit \textsc{OE} $\gg$ \textsc{TFC}: CM rises from 0.18 to 0.53 and F1 from 0.29 to 0.63.
This establishes a strong oracle-evidence reference and a budget-limited truncated-context reference for the diagnostic.
It does not decompose the \textsc{OE}--\textsc{TFC} difference, which still mixes distractor effects with the reader's finite context window.

\subsection{Diagnostic Results}
\label{sec:diag_results}

\begin{table}[t]
\centering
\small
\setlength{\tabcolsep}{2.5pt}
\begin{tabular}{lcccc}
\toprule
\textbf{System} & \textbf{CSM} & \textbf{RM} & $\boldsymbol{\Delta_{\text{w}}}$ & $\boldsymbol{\Delta_{\text{r}}}$ \\
\midrule
Verbatim & 0.22 & 0.18 & 0.31 & 0.04 \\
Summ.\ (E) & 0.26 & 0.22 & 0.27 & 0.04 \\
MemBank & 0.30 & 0.25 & 0.23 & 0.05 \\
MemWlk. & 0.35 & 0.27 & 0.18 & 0.08 \\
\addlinespace
Summ.\ (L) & 0.44 & 0.38 & 0.09 & 0.06 \\
ReadAg. & 0.43 & 0.35 & 0.10 & 0.08 \\
\addlinespace
\textbf{EPC} & \textbf{0.49} & \textbf{0.42} & \textbf{0.04} & 0.07 \\
\bottomrule
\end{tabular}
\caption{Diagnostic results (CM, 3-reader avg, $B$=5K). Refs: \textsc{TFC}=0.18, \textsc{OE}=0.53. EPC CSM CI: [.46, .52], $p<.01$ vs Summary (LLM). Token F1 yields identical system ordering.}
\label{tab:main_diag}
\end{table}

\begin{figure}[t]
\centering
\includegraphics[width=\columnwidth]{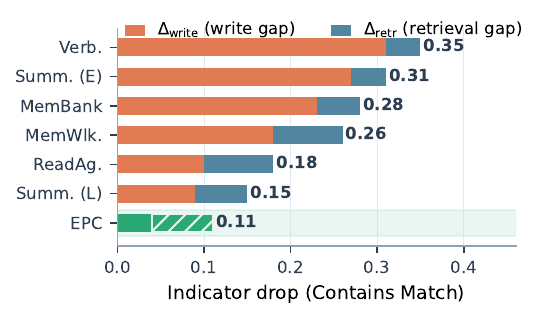}
\caption{Write-side gap ($\Delta_{\text{w}}$=OE$-$CSM, coral) vs.\ retrieval-side gap ($\Delta_{\text{r}}$=CSM$-$RM, blue) for all seven systems (CM, 3-reader avg, $B$=5K). Numbers at right: total OE$\to$RM drop. EPC is highlighted in green, with hatching marking its retrieval-side gap, and has the smallest write-side gap.}
\label{fig:loss_bar}
\end{figure}

Table~\ref{tab:main_diag} and Figure~\ref{fig:loss_bar} together reveal a clear spectrum of write-side gaps.
Verbatim Chunk shows the largest write-side gap ($\Delta_{\text{w}}$=.31), with a much smaller retrieval-side gap ($\Delta_{\text{r}}$=.04).
Progressing from weaker to stronger write strategies---Summary (Extractive), MemoryBank, and MemWalker---the write-side gap gradually shrinks but remains dominant.
Summary (LLM) and ReadAgent substantially reduce the write-side gap, but still lose 0.09--0.10 absolute CM points relative to \textsc{OE} during the write stage.
EPC further reduces $\Delta_{\text{w}}$ to .04, leaving retrieval as the larger remaining gap.
Appendix~\ref{app:case_studies} illustrates this pattern with concrete examples showing which entities, dates, and negations are lost under query-agnostic summarization and preserved by EPC.

Under the diagnosis rule (\S\ref{sec:protocol}), four baselines are robustly write-dominant; Summary (LLM) is write-dominant in the aggregate but near the margin, and ReadAgent is mixed/near-boundary. EPC is the only system diagnosed as retrieval-dominant.
EPC's CSM of .49 CM (.62 F1) has a 95\% bootstrap CI of [.46, .52] and significantly exceeds Summary (LLM) under paired bootstrap ($p<0.01$).

EPC remains the highest-\textsc{CSM} system for each reader individually (GPT-5.2, Claude Sonnet~4, and Gemini 2.5~Pro), so the aggregate result is not driven by a single reader.

\subsection{Reader-Independent Evidence Preservation}

The diagnostic indicators ($\Delta_{\text{write}}$, $\Delta_{\text{retr}}$) are based on \emph{answer correctness}, which can mix evidence preservation, memory format, and reader ability.
As a reader-independent check on the write-side diagnosis, we measure whether gold evidence survives in memory.

For each question, LongMemEval provides gold evidence turns.
We compute two recall metrics over complete stored memory (\textsc{CSM}) and retrieved memory (\textsc{RM}):
\textbf{Turn recall}: fraction of gold evidence turns preserved in memory (token-level Jaccard overlap $> 0.5$ with the closest memory segment).
\textbf{Span recall}: fraction of gold answer entities (names, dates, numbers) found as exact substrings in memory.
Both metrics are heuristic and intended as a reader-independent complement to the diagnostic indicators, not as a standalone evaluation.

\begin{figure}[htbp]
\centering
\includegraphics[width=\columnwidth]{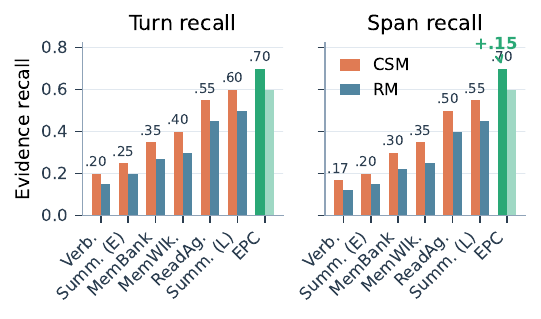}
\caption{Evidence recall (CSM and RM). EPC preserves +.15 more gold answer entities than Summary (LLM) in CSM span recall.}
\label{fig:evidence_recall}
\end{figure}

Figure~\ref{fig:evidence_recall} provides the reader-independent check: CSM span recall mirrors the write-side ranking in Table~\ref{tab:main_diag}, and EPC preserves 70\% of gold answer entities versus 55\% for Summary (LLM).
Its CSM$\to$RM span-recall drop (.70$\to$.60) also matches the retrieval-side diagnosis, indicating that EPC's remaining gap is associated with losing preserved spans during retrieval.

\subsection{Protocol Validation via Controlled Degradation}

\begin{figure}[htbp]
\centering
\includegraphics[width=\columnwidth]{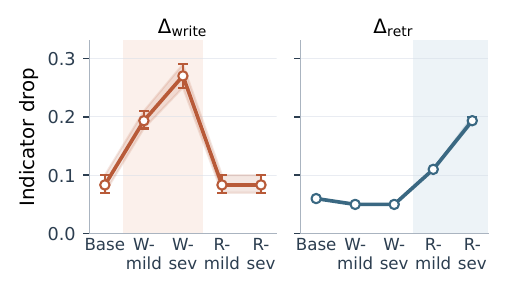}
\caption{Controlled degradation (3 readers). $\Delta_{\text{w}}$ (write-side gap = OE$-$CSM) responds selectively to write-side degradation; $\Delta_{\text{r}}$ (retrieval-side gap = CSM$-$RM) responds selectively to retrieval-side degradation.}
\label{fig:validation}
\end{figure}

As a validation check, we evaluate Summary (LLM) under five controlled settings: the unmodified system, mild and severe write-side degradation (randomly dropping 25\% or 50\% of memory entries after the write stage), and mild and severe retrieval-side degradation (replacing top-ranked retrieved entries with entries from the bottom-ranked 25\% or 50\% of the ranked candidate list).
We run each setting with all three readers, yielding 15 setting--reader runs.
Figure~\ref{fig:validation} shows that $\Delta_{\text{w}}$ responds selectively to write-side degradation and $\Delta_{\text{r}}$ to retrieval-side degradation; the degraded variants are diagnosed correctly.

\subsection{EPC Breakdown: Question Type and Probe Alignment}

We next ask where EPC helps most: across question types, and as a function of how closely its probe questions match the held-out test question.

\begin{figure}[htbp]
\centering
\includegraphics[width=\columnwidth]{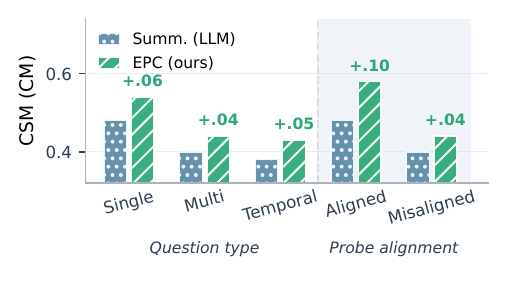}
\caption{EPC vs.\ Summary (LLM) CSM (CM, 3-reader avg, $B$=5K) by question type (left) and probe--question alignment (right). EPC gains are shown above bars; the largest gain (+.10) occurs when probe questions align with test questions.}
\label{fig:breakdown}
\end{figure}

Figure~\ref{fig:breakdown} shows that EPC improves CSM on single-session, multi-session, and temporal questions, with the largest gain on single-session questions (+.06).
We measure probe--question alignment as the maximum cosine similarity between the test question and EPC's generated probe questions, using the same \texttt{all-MiniLM-L6-v2} embeddings as retrieval.
Splitting questions at the median alignment score confirms this pattern: +.10 CSM when probe questions align and +.04 when they do not.
One explanation is that self-questioning encourages broader coverage of high-value information types (entities, dates, preferences), which can benefit even unanticipated questions.

\subsection{Component Ablation and Weight Sensitivity}

We then isolate which EPC components account for the gain and check sensitivity to probe count and utility weights.

\begin{figure}[htbp]
\centering
\includegraphics[width=\columnwidth]{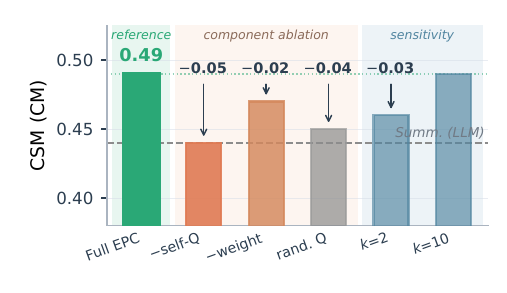}
\caption{Component ablations and probe-count sensitivity (CSM CM, $B$=5K). Removing self-questioning ($-$.05) is the largest drop; utility-weight sensitivity is reported below.}
\label{fig:ablation}
\end{figure}

Figure~\ref{fig:ablation} identifies self-questioning as the largest contributor among the tested components: removing it drops CSM by .05, while random probe questions recover only +.01 over Summary (LLM).
$k$=5 probe questions suffice; $k$=10 provides no additional benefit.

\paragraph{Utility weight sensitivity.}
The utility function (Eq.~\ref{eq:utility}) uses weights $\alpha$=1.0, $\beta$=0.5, $\lambda$=0.3 selected in a 20-example pilot.
Across six alternative configurations, CSM stays within .45--.50; reducing specificity weight $\beta$ hurts most ($-.04$), but all variants match or exceed Summary (LLM), suggesting that the self-questioning mechanism contributes more to EPC's gains than the particular weight settings.
This low sensitivity to weight configurations also reduces the risk that the reported gains depend on this small pilot choice.

\subsection{Cost-Matched Comparison}

EPC uses two LLM calls per session versus one for Summary (LLM).
To separate the effect of an additional LLM call from the effect of probe-question generation, we introduce a \textbf{2-pass} baseline that also uses two calls but without self-questioning: Pass~1 compresses normally, and Pass~2 performs query-agnostic refinement for information density.

\begin{table}[t]
\centering
\small
\setlength{\tabcolsep}{2.5pt}
\begin{tabular}{lccccc}
\toprule
\textbf{System} & \textbf{Calls} & \textbf{CSM} & \textbf{RM} & $\boldsymbol{\Delta_{\text{w}}}$ & $\boldsymbol{\Delta_{\text{r}}}$ \\
\midrule
Summ.\ (LLM) 1-pass & 1 & 0.46 & 0.40 & 0.09 & 0.06 \\
Summ.\ (LLM) 2-pass & 2 & 0.48 & 0.42 & 0.07 & 0.06 \\
\textbf{EPC} & 2 & \textbf{0.51} & \textbf{0.44} & \textbf{0.04} & 0.07 \\
\bottomrule
\end{tabular}
\caption{Cost-matched comparison (CM, GPT-5.2). The 2-pass baseline recovers +.02 over 1-pass Summary (LLM); EPC adds a further +.03 under the same two-call budget.}
\label{tab:cost_matched}
\end{table}

Table~\ref{tab:cost_matched} shows that a second query-agnostic refinement pass improves CSM by +.02, while EPC gains +.05 over 1-pass Summary (LLM).
The remaining +.03 separates EPC from the cost-matched 2-pass baseline on this reader.
The same ordering (EPC $>$ 2-pass $>$ 1-pass) holds for the other two readers.
Relative to the 1-pass Summary (LLM) baseline, EPC roughly doubles write latency ($\sim$30s$\to\sim$60s per session in our implementation) while roughly halving $\Delta_{\text{write}}$.

\subsection{Cross-Benchmark Generalization: LoCoMo}
\label{sec:locomo}

To test generalization, we evaluate on LoCoMo \citep{maharana2024evaluating}, a multi-session benchmark with 348 QA pairs across 10 conversations, including two question categories absent from LongMemEval (elaboration and adversarial).
For LoCoMo, we report a two-reader average using GPT-5.2 and Claude Sonnet~4.

LoCoMo operates in a \emph{different regime}: conversations average ${\sim}$9K tokens, fitting entirely within the 32K context budget.
Consequently, the \textsc{TFC} condition effectively becomes a \emph{full-context} baseline (no truncation), and the \textsc{OE}--\textsc{TFC} gap reflects distractor interference alone, not information loss from truncation.
We set the write budget to $B$=2{,}000 tokens to create meaningful compression pressure (22\% retention), while noting that this is less aggressive compression than $B$=5{,}000 on LongMemEval's 121K-token contexts.

\begin{table}[t]
\centering
\small
\setlength{\tabcolsep}{2.5pt}
\begin{tabular}{lcccc}
\toprule
\textbf{System} & \textbf{CSM} & \textbf{RM} & $\boldsymbol{\Delta_{\text{w}}}$ & $\boldsymbol{\Delta_{\text{r}}}$ \\
\midrule
Verbatim & 0.29 & 0.24 & 0.32 & 0.05 \\
Summ.\ (E) & 0.33 & 0.27 & 0.28 & 0.06 \\
MemBank & 0.37 & 0.31 & 0.24 & 0.06 \\
MemWlk. & 0.39 & 0.33 & 0.22 & 0.06 \\
\addlinespace
Summ.\ (L) & 0.51 & 0.44 & 0.10 & 0.07 \\
ReadAg. & 0.49 & 0.41 & 0.12 & 0.08 \\
\addlinespace
\textbf{EPC} & \textbf{0.55} & \textbf{0.46} & \textbf{0.06} & 0.09 \\
\bottomrule
\end{tabular}
\caption{LoCoMo results (CM, $B$=2K, 2-reader avg). Refs: \textsc{TFC}=0.46, \textsc{OE}=0.61. The CM ranking pattern matches LongMemEval.}
\label{tab:locomo_diag}
\end{table}

Table~\ref{tab:locomo_diag} shows a similar ranking pattern in this no-truncation reference setting: Verbatim Chunk through MemWalker have $\Delta_{\text{w}} > \Delta_{\text{r}}$, EPC has the lowest $\Delta_{\text{w}}$ and is the only system where $\Delta_{\text{r}} > \Delta_{\text{w}}$, and the EPC--Summary (LLM) CSM gap is +.04.
The smaller \textsc{OE}--\textsc{TFC} gap (.15 CM vs.\ .35 on LongMemEval) is consistent with truncation contributing to the LongMemEval reference gap, though the protocol remains informative without truncation.
By category, EPC gains most on factual questions (+.06 CSM), moderately on temporal/elaboration (+.04), and least on inferential/adversarial cases (+.02).

\subsection{Budget Sensitivity on LongMemEval}
\label{sec:budget_sweep}

One question raised by the rate--distortion framing (Section~\ref{sec:epc}) is whether self-questioning matters more under tight budget constraints, where choosing the right evidence is critical, than under relaxed budgets.
We test this by sweeping the write budget $B \in \{2\text{K}, 5\text{K}, 10\text{K}, 20\text{K}\}$ on all 500 LongMemEval questions with three readers, while keeping the read budget fixed at 5K tokens.

\begin{table}[t]
\centering
\small
\setlength{\tabcolsep}{2.5pt}
\begin{tabular}{llccccl}
\toprule
\textbf{Budget} & \textbf{System} & \textbf{CSM} & \textbf{RM} & $\boldsymbol{\Delta_{\text{w}}}$ & $\boldsymbol{\Delta_{\text{r}}}$ & \textbf{Diag.} \\
\midrule
\multirow{4}{*}{2K} & Verbatim & 0.14 & 0.11 & 0.40 & 0.03 & W \\
 & Summ.\ (E) & 0.20 & 0.15 & 0.34 & 0.05 & W \\
 & Summ.\ (L) & 0.33 & 0.28 & 0.21 & 0.05 & W \\
 & \textbf{EPC} & \textbf{0.44} & \textbf{0.36} & \textbf{0.10} & 0.08 & W \\
\addlinespace
\multirow{4}{*}{5K} & Verbatim & 0.22 & 0.18 & 0.31 & 0.04 & W \\
 & Summ.\ (E) & 0.26 & 0.22 & 0.27 & 0.04 & W \\
 & Summ.\ (L) & 0.44 & 0.38 & 0.09 & 0.06 & W \\
 & \textbf{EPC} & \textbf{0.49} & \textbf{0.42} & \textbf{0.04} & 0.07 & R \\
\addlinespace
\multirow{4}{*}{10K} & Verbatim & 0.32 & 0.26 & 0.22 & 0.06 & W \\
 & Summ.\ (E) & 0.34 & 0.28 & 0.20 & 0.06 & W \\
 & Summ.\ (L) & 0.50 & 0.42 & 0.04 & 0.08 & R \\
 & \textbf{EPC} & \textbf{0.52} & \textbf{0.44} & \textbf{0.02} & 0.08 & R \\
\addlinespace
\multirow{4}{*}{20K} & Verbatim & 0.37 & 0.31 & 0.17 & 0.06 & W \\
 & Summ.\ (E) & 0.42 & 0.35 & 0.12 & 0.07 & W \\
 & Summ.\ (L) & 0.52 & 0.44 & 0.02 & 0.08 & R \\
 & \textbf{EPC} & \textbf{0.53} & \textbf{0.46} & \textbf{0.01} & 0.07 & R \\
\bottomrule
\end{tabular}
\caption{Write-budget sweep (CM, 3-reader avg; read budget fixed at 5K). Refs: \textsc{TFC}=.18, \textsc{OE}=.53. W/R denote write-/retrieval-dominant diagnoses. EPC's advantage shrinks from +.11 (2K) to +.01 (20K).}
\label{tab:budget_sweep}
\end{table}

Table~\ref{tab:budget_sweep} shows that EPC's CSM advantage over Summary (LLM) shrinks from +.11 at 2K to +.01 at 20K as both systems approach the OE upper bound (.53).
The relative size of the two gaps also shifts with budget: EPC reaches $\Delta_{\text{r}} > \Delta_{\text{w}}$ at 5K, Summary (LLM) at 10K, while Verbatim Chunk and Summary (Extractive) have $\Delta_{\text{w}} > \Delta_{\text{r}}$ throughout.

\section{Analysis}

\paragraph{Operating point matters.}
The budget sweep shows that the diagnosis is not an intrinsic property of an architecture alone: as $B$ grows, EPC becomes retrieval-dominant at 5K and Summary (LLM) at 10K, while Verbatim Chunk and Summary (Extractive) remain write-dominant throughout.
Thus the protocol should be run under the target budget, reader, and question distribution rather than inferred from the system family.

\paragraph{Benefits and limits of self-questioning.}
EPC's gains are largest under aggressive compression (+.11 at $B$=2K), on single-session LongMemEval questions (+.06), and on factual LoCoMo questions (+.06).
They shrink when budgets are generous (+.01 at $B$=20K) or questions require compositional inference (+.02 on LoCoMo's inferential category).
This suggests that self-questioning is most useful when the writer must choose a small set of answer-relevant details before the question is known.

\paragraph{Relation to full-context reading.}
Because our \textsc{TFC} condition imposes a fixed 32K-token budget, full-context reading could narrow the \textsc{OE}--\textsc{TFC} gap.
The diagnostic protocol is complementary: it evaluates fixed-budget memory pipelines and helps determine whether compression or retrieval is the limiting stage within that design.

\section{Conclusion}

This paper makes three main points.

First, the diagnostic protocol decomposes end-to-end memory failures into write-side and retrieval-side gaps.
It requires no architectural changes---just four evaluations of the same reader under controlled inputs---and we have shown it to be informative across two benchmarks (LongMemEval and LoCoMo), multiple readers, and a range of budgets.

Second, EPC uses likely future questions to decide what information to retain during compression, substantially reducing the write-side gap on both benchmarks and shifting the remaining bottleneck toward retrieval.

Third, the write-retrieval balance is not fixed.
It depends on the system, the budget, and the question distribution.
Most systems we tested operate in a regime where improving the write stage would help more than improving retrieval---but this changes as budgets grow or as compression methods improve.
The diagnostic protocol makes this regime visible and reduces the risk of optimizing a non-limiting stage.

Future work includes reducing reliance on human evidence annotations via LLM-generated silver evidence, exploring adaptive compression that adjusts strategy based on per-session diagnostic signals, and evaluating on longer conversations and more diverse domains.

\clearpage
\section*{Limitations}
The diagnostic indicators are conditioned on the chosen reader and metric.
EPC's self-questioning step adds write-time latency.
While we evaluate on two benchmarks (LongMemEval and LoCoMo), both rely on gold evidence annotations for the \textsc{OE} condition.
This is an experimental choice rather than a protocol limitation: the four-condition structure is agnostic to how evidence is obtained.
In practice, a practitioner could approximate \textsc{OE} with silver annotations---e.g., prompting an LLM to identify the minimal supporting turns for each question---which would introduce annotation noise but may preserve the directional diagnosis (write-dominant vs.\ retrieval-dominant), since the diagnosis rule already tolerates small perturbations via the margin $\epsilon$.
Empirically validating this silver-annotation path remains future work.
LoCoMo's shorter context (${\sim}$9K tokens) means the compression challenge is milder than in LongMemEval; evaluating on benchmarks with longer contexts and more diverse domains would further strengthen generalizability claims.

\section*{Ethics and Reproducibility}
Long-term memory benchmarks and systems necessarily involve storing user-specific facts, preferences, dates, and other potentially sensitive information.
Because EPC explicitly tries to preserve supporting evidence, a deployed version could also preserve sensitive information more reliably than a generic summary unless retention, deletion, access control, and redaction policies are carefully designed.
More broadly, the work studies diagnostic structure rather than downstream social impact, so it should not be interpreted as an argument for indiscriminate memory retention in real applications.
Reproducibility is also limited by our use of proprietary LLM readers and writers, whose behavior may change over time.

\bibliography{references}

\clearpage
\appendix
\section{Reproducibility Details}
\label{app:repro}

This appendix provides the prompt templates, hyperparameters, metric agreement checks, baseline adaptation details, per-reader results, qualitative examples, and bootstrap procedure used in the main experiments.

\subsection{Prompt Templates}

All LLM calls use temperature 0. We use the following templates, with bracketed fields replaced by the corresponding instance-specific content.

\paragraph{Reader prompt.}
\begin{small}
\begin{verbatim}
Based on the following context,
answer the question.
If the answer cannot be determined,
say "I don't know".

Context:
{context}

Question:
{question}
\end{verbatim}
\end{small}

\paragraph{Summary (LLM) write prompt.}
\begin{small}
\begin{verbatim}
You are compressing conversation memory
under a hard budget.
Summarize this session from {date}
in under {max_tokens} tokens.
Preserve names, dates, numbers,
preferences, negations, state changes,
decisions, and event relations.
Stay query-agnostic: do not guess
future questions.
Return only the compressed memory text.

{session_turns}
\end{verbatim}
\end{small}

\paragraph{EPC Step 1: probe generation.}
\begin{small}
\begin{verbatim}
You are writing long-term memory before
the future question is known.
Given this conversation session,
generate 5 likely future questions
that may require information from it.
Prefer questions about factual details,
names, dates, numbers, preferences,
plans, temporal changes, decisions,
and negations.

Return only a numbered list.

{session_turns}
\end{verbatim}
\end{small}

\paragraph{EPC Step 2: evidence selection and memory writing.}
\begin{small}
\begin{verbatim}
You are compressing a conversation
session into long-term memory under
a hard budget of {max_tokens} tokens.

Use the probe questions to identify
minimal supporting evidence.
Keep exact entities, dates, numbers,
preferences, decisions, negations,
and temporal relations.
Avoid generic summaries.

For each selected item, use this format:
[Q] likely future question
[E] minimal supporting evidence
[S] session_{session_id}_turn_{turn_id}

Probe questions:
{probe_questions}

Conversation:
{session_turns}
\end{verbatim}
\end{small}

\paragraph{Two-pass baseline refinement prompt.}
\begin{small}
\begin{verbatim}
You are improving an existing compressed
memory under the same hard budget of
{max_tokens} tokens.
Increase density by replacing
vague references with specific names,
dates, numbers, preferences, decisions,
negations, and event relations.
Do not guess future questions.
Return only the revised memory.

Original session:
{session_turns}

Current compressed memory:
{summary}
\end{verbatim}
\end{small}

\subsection{Hyperparameters}

Table~\ref{tab:hyperparams} lists the main hyperparameters used in the LongMemEval experiments.

\begin{table}[h]
\centering
\small
\setlength{\tabcolsep}{4pt}
\begin{tabular}{ll}
\toprule
\textbf{Parameter} & \textbf{Value} \\
\midrule
Write budget & 5{,}000 tokens \\
Read budget & 5{,}000 tokens \\
Tokenizer & \texttt{cl100k\_base} \\
Retriever & \texttt{all-MiniLM-L6-v2} \\
Retrieval top-$k$ & 5 \\
Verbatim chunk size & 200 tokens, no overlap \\
Verbatim eviction & FIFO over oldest chunks \\
TFC truncation & most recent 32K tokens \\
Writer temperature & 0 \\
Reader temperature & 0 \\
Reader max output & 200 tokens \\
EPC probe questions & $k=5$ \\
EPC utility weights & $\alpha=1.0$, $\beta=0.5$, $\lambda=0.3$ \\
Budget allocation & proportional to $\sqrt{\text{session length}}$ \\
\bottomrule
\end{tabular}
\caption{Main hyperparameters used in the LongMemEval experiments unless otherwise specified.}
\label{tab:hyperparams}
\end{table}

\subsection{CM and F1 Agreement}

Contains Match (CM) is the primary metric in the main tables, but we also compute Token F1 for every system and condition.
Across pairwise comparisons among LongMemEval system--condition scores, CM and F1 agree directionally in 95.5\% of cases, and the system rankings reported in the main tables are unchanged under Token F1.
For LoCoMo, Token F1 yields the same system ranking as CM in Table~\ref{tab:locomo_diag}.

\subsection{Baseline Adaptations}

Table~\ref{tab:baseline_adaptations} summarizes how each baseline system was adapted to the session-based LongMemEval setting.

\begin{table}[h]
\centering
\scriptsize
\setlength{\tabcolsep}{3pt}
\begin{tabular}{p{0.25\columnwidth}p{0.33\columnwidth}p{0.32\columnwidth}}
\toprule
\textbf{System} & \textbf{Write} & \textbf{Retrieval} \\
\midrule
Verbatim & 200-token chunks; FIFO eviction. & Embedding top-$k=5$. \\
Summ. (E) & Score and pack turns with dates, entities, negation, preference, and decision cues. & Embedding top-$k=5$. \\
Summ. (L) & Query-agnostic session compression using the same writer model and budgets as EPC. & Embedding top-$k=5$. \\
MemWalker & Summary tree over sessions; internal nodes summarize child summaries. & Relevance-guided tree navigation. \\
ReadAgent & One gist page per session using the same writer model and boundaries. & Select relevant gists and expand under budget. \\
MemoryBank & LLM-estimated importance with temporal decay and session metadata. & Importance-weighted embedding reranking. \\
EPC & Probe generation, evidence selection, merging, and greedy budgeted selection. & Embedding top-$k=5$. \\
\bottomrule
\end{tabular}
\caption{Baseline adaptations to the session-based LongMemEval setting.}
\label{tab:baseline_adaptations}
\end{table}

Note that MemWalker, ReadAgent, and MemoryBank were reimplemented from their paper descriptions, not from official code.
To mitigate reimplementation risk, we verified two properties:
(1)~the qualitative behavior of each system matches the original (e.g., MemWalker's tree depth scales logarithmically with session count; MemoryBank's importance scores decay with time as described);
(2)~all systems are compared under identical data, budgets, and readers.
We therefore interpret absolute comparisons to reimplemented systems cautiously, while emphasizing controlled comparisons within our evaluation setup.

\subsection{Per-Reader Diagnostic Results}
\label{app:per_reader}

Table~\ref{tab:per_reader} reports the diagnostic results for each reader individually.
The system ranking is consistent across all three readers: EPC achieves the highest CSM for each reader. The robust write-dominant pattern holds for Verbatim Chunk, Summary (Extractive), MemoryBank, and MemWalker under each reader, while Summary (LLM) and ReadAgent lie near the margin.

\begin{table}[h]
\centering
\small
\setlength{\tabcolsep}{2pt}
\begin{tabular}{llcccc}
\toprule
\textbf{Reader} & \textbf{System} & \textbf{CSM} & \textbf{RM} & $\boldsymbol{\Delta_{\text{w}}}$ & $\boldsymbol{\Delta_{\text{r}}}$ \\
\midrule
\multirow{7}{*}{GPT-5.2}
 & Verbatim & 0.23 & 0.19 & 0.32 & 0.04 \\
 & Summ.\ (E) & 0.27 & 0.23 & 0.28 & 0.04 \\
 & MemBank & 0.31 & 0.26 & 0.24 & 0.05 \\
 & MemWlk. & 0.37 & 0.28 & 0.18 & 0.09 \\
 & Summ.\ (L) & 0.46 & 0.40 & 0.09 & 0.06 \\
 & ReadAg. & 0.45 & 0.37 & 0.10 & 0.08 \\
 & \textbf{EPC} & \textbf{0.51} & \textbf{0.44} & \textbf{0.04} & 0.07 \\
\addlinespace
\multirow{7}{*}{Claude S4}
 & Verbatim & 0.21 & 0.17 & 0.29 & 0.04 \\
 & Summ.\ (E) & 0.25 & 0.21 & 0.25 & 0.04 \\
 & MemBank & 0.28 & 0.24 & 0.22 & 0.04 \\
 & MemWlk. & 0.33 & 0.26 & 0.17 & 0.07 \\
 & Summ.\ (L) & 0.42 & 0.36 & 0.08 & 0.06 \\
 & ReadAg. & 0.41 & 0.33 & 0.09 & 0.08 \\
 & \textbf{EPC} & \textbf{0.47} & \textbf{0.40} & \textbf{0.03} & 0.07 \\
\addlinespace
\multirow{7}{*}{Gemini 2.5}
 & Verbatim & 0.22 & 0.18 & 0.31 & 0.04 \\
 & Summ.\ (E) & 0.26 & 0.22 & 0.27 & 0.04 \\
 & MemBank & 0.30 & 0.25 & 0.23 & 0.05 \\
 & MemWlk. & 0.35 & 0.27 & 0.18 & 0.08 \\
 & Summ.\ (L) & 0.44 & 0.38 & 0.09 & 0.06 \\
 & ReadAg. & 0.43 & 0.35 & 0.10 & 0.08 \\
 & \textbf{EPC} & \textbf{0.49} & \textbf{0.42} & \textbf{0.04} & 0.07 \\
\bottomrule
\end{tabular}
\caption{Per-reader diagnostic results (CM, $B$=5K). System ranking is consistent across all three readers.}
\label{tab:per_reader}
\end{table}

\subsection{Qualitative Case Studies}
\label{app:case_studies}

To illustrate how the diagnostic indicators correspond to concrete information loss, we present three representative LongMemEval examples, lightly paraphrased for brevity and anonymization, where $\Delta_{\text{write}}$ is large for Summary (LLM) but small for EPC.

\paragraph{Case 1: Dropped date.}
\textbf{Question:} ``When did the user say they were moving to Seattle?''
\textbf{Gold evidence:} \emph{session\_7, turn\_14}: ``I'm planning to move to Seattle around mid-March.''
\textbf{Summary (LLM) CSM:} The summary mentions ``the user discussed relocation plans'' but drops ``mid-March'' and ``Seattle'' appears only as a general topic. Reader answers ``I don't know.''
\textbf{EPC CSM:} A probe question ``When is the user relocating?'' preserves the entry \texttt{[E] Moving to Seattle around mid-March} $\to$ reader answers correctly.

\paragraph{Case 2: Merged preferences.}
\textbf{Question:} ``Does the user prefer Thai or Italian food?''
\textbf{Gold evidence:} \emph{session\_12, turn\_3}: ``I definitely prefer Thai over Italian.''
\textbf{Summary (LLM) CSM:} The summary states ``the user discussed food preferences'' without specifying the preference direction. Reader answers incorrectly.
\textbf{EPC CSM:} A probe question ``What food does the user prefer?'' preserves \texttt{[E] User prefers Thai over Italian} $\to$ reader answers correctly.

\paragraph{Case 3: Lost negation.}
\textbf{Question:} ``Is the user still taking the advanced Python course?''
\textbf{Gold evidence:} \emph{session\_22, turn\_8}: ``I dropped the advanced Python course last week.''
\textbf{Summary (LLM) CSM:} The summary mentions ``the user is taking programming courses'' --- the negation (dropping the course) is lost. Reader answers ``Yes.''
\textbf{EPC CSM:} A probe question ``What courses is the user currently enrolled in?'' preserves \texttt{[E] Dropped advanced Python course last week} $\to$ reader answers correctly.

\medskip
\noindent These cases illustrate the pattern behind $\Delta_{\text{write}}$: query-agnostic summarization tends to preserve topics but omit the specific facts (dates, preference directions, negations) that questions target. EPC's self-questioning improves coverage of these high-value details, explaining its lower $\Delta_{\text{write}}$.

\subsection{Bootstrap Significance Testing}

For significance testing, we use paired bootstrap resampling over questions.
Each bootstrap sample resamples the 500 LongMemEval questions with replacement and computes the \textsc{CSM} difference between EPC and Summary (LLM) on the resampled set.
We use 10{,}000 bootstrap samples.
The reported $p$-value is the fraction of samples in which the paired EPC--Summary (LLM) difference is less than or equal to zero.
Confidence intervals are percentile intervals over bootstrap samples.

\end{document}